\documentclass[12pt,a4paper,twoside]{article}
\usepackage[]{graphicx}
\usepackage{pstricks,pst-text,pst-node,pst-plot}
\usepackage{epsf}
\usepackage{amssymb}
\usepackage{amsmath}
\newtheorem{theorem}{Theorem}
\newtheorem{corollary}[theorem]{Corollary}
\newtheorem{definition}[theorem]{Definition}

\newenvironment{proof}[1][Proof]{\textbf{#1.} }{\
\rule{0.5em}{0.5em}}

\newcommand{\adj}{{\rm adj}}
\newcommand{\card}{{\rm card}}
\newcommand{\supp}{{\rm supp}}

\title{A Class of LULU Operators on\\Multi-Dimensional Arrays}
\author{Roumen Anguelov and Inger Plaskitt\\
Department of Mathematics and Applied Mathematics\\
University of
Pretoria\\roumen.anguelov@up.ac.za\\inger.plaskitt@up.ac.za}

\date{}

\begin{document}

\maketitle

\abstract{The LULU operators for sequences are extended to
multi-dimensional arrays via the morphological concept of
connection in a way which preserves their essential properties,
e.g. they are separators and form a four element fully ordered
semi-group. The power of the operators is demonstrated by deriving
a total variation preserving discrete pulse decomposition of
images.}

\vspace{12pt}

\noindent\emph{Keywords: LULU, connection, separator, discrete
pulse transform, total variation.}

\section{Introduction}

The LULU operators and the associated Discrete Pulse Transform
developed during the last two decades or so are an important
contribution to the theory of the nonlinear multi-resolution
analysis of sequences. The basics of the theory as well as the
most significant results until 2005 are published in the monograph
\cite{Rohwerbook}. For more recent developments and applications
see \cite{Anguelov}, \cite{Conradie}, \cite{Kao}, \cite{Laurie},
\cite{RohwerWild}. Central to the theory is the concept of
\emph{separator}. This concept is defined in \cite{Rohwerbook}
only for operators on sequences due to the context of the book.
However, it is meaningful in more general settings. In fact, some
of the axioms have been used earlier, e.g. see \cite{Serra82}, for
functions on arbitrary domains. We will give the definition of
separator for operators on real functions defined on a domain with
a group structure.

Let a $\Omega$ be an abelian group. Denote by
$\mathcal{A}(\Omega)$ the vector lattice of all real functions
defined on $\Omega$ with respect to the usual point-wise defined
addition, scalar multiplication and partial order. For every
$a\in\Omega$ the operator $E_a:\mathcal{A}(\Omega)\to
\mathcal{A}(\Omega)$ given by
\[
E_a(f)(x)=f(x+a),\ x\in\Omega,
\]
is called a shift operator.
\begin{definition}\label{defSep}
An operator $P:\mathcal{A}(\Omega)\to \mathcal{A}(\Omega)$ is
called a {\bf separator} if
\begin{eqnarray*}
&(i)&P\circ E_a=E_a\circ P,\ a\in\Omega;\hspace{10cm}\\
&(ii)&P(f+c)=P(f)+c,\ f,c\in\mathcal{A}(\Omega), \ c\mbox{ -
constant function};\\
&(iii)&P(\alpha f)=\alpha P(f),\ \alpha\in\mathbb{R},\ \alpha\geq
0, \ f\in\mathcal{A}(\Omega);\\
&(iv)&P\circ P=P;\hspace{7.2cm}\mbox{(Idempotence)}\\
&(v)&(id-P)\circ (id-P)=id-P.\hspace{4cm}\mbox{(Co-idempotence)}
\end{eqnarray*}
\end{definition}
Here $id$ denotes the identity operator and the operator $id-P$ is
defined in terms of the point-wise linear operations for the
operators on $\mathcal{A}(\Omega)$, that is, $(id-P)(f)=f-P(f)$.
The first two axioms in Definition \ref{defSep} and partially the
third one were first introduced as required properties of
nonlinear smoothers by Mallows, \cite{Mallows}. Rohwer further
made the concept of a smoother more precise by using the
properties (i)--(iii) as a definition of this concept. The axiom
(iv) is an essential requirement for what is called a {\it
morphological filter}, \cite{Serra82}, \cite{Serra88},
\cite{Soille}. In fact, a morphological filter is exactly a
syntone operator which satisfies (iv). Let us recall that an
operator $P$ is called \emph{syntone} if
\[
f\leq g \Longrightarrow P(f)\leq P(g)\ ,\ \
f,g\in\mathcal{A}(\Omega).
\]
The co-idempotence axiom (v) in Definition \ref{defSep} was
introduced by Rohwer in \cite{Rohwerbook}, where it is also shown
that it is an essential requirement for operators extracting
signal from a sequence.

The LULU theory was developed for sequences, that is, the case
$\Omega=\mathbb{Z}$. Given a bi-infinite sequence
$\xi=(\xi_i)_{i\in\mathbb{Z}}$ and $n\in\mathbb{N}$ the basic LULU
operators $L_n$ and $U_n$ are defined as follows
\begin{eqnarray}
(L_n
\xi)_i=\max\{\min\{\xi_{i-n},...,\xi_i\},...,\min\{\xi_{i},...,\xi_{i+n}\}\},\
i\in\mathbb{Z}.\label{LnSec}\\
(U_n
\xi)_i=\min\{\max\{\xi_{i-n},...,\xi_i\},...,\max\{\xi_{i},...,\xi_{i+n}\}\},\
i\in\mathbb{Z}.\label{UnSec}
\end{eqnarray}
It is shown in \cite{Rohwerbook} that for every $n\in\mathbb{N}$
the operators $L_n$ and $U_n$ as well as their compositions are
syntone separators. Hence they are an appropriate tool for signal
extraction. Furthermore, these operators form the so called strong
LULU semi-group. This a four element semi-group with respect to
composition, see Table 1, which is fully ordered with respect to
the usual point-wise defined order
\begin{equation}\label{defOrder}
P\leq Q\Longleftrightarrow P(f)\leq Q(f),\
f\in\mathcal{A}(\Omega).
\end{equation}
We have%
\begin{equation}\label{LULUorder}
L_n\leq U_n\circ L_n\leq L_n\circ U_n\leq U_n.
\end{equation}

\begin{table}[h]
\begin{center}
\begin{tabular}{c||c|c|c|c|}
&$ L_n$&$ U_n$&$U_n\circ L_n$&$L_n\circ U_n$\\\hline\hline $L_n$&$
L_n$&$
 L_n\circ U_n$&$U_n\circ L_n$
&$ L_n\circ U_n$\\\hline $ U_n$&$ U_n\circ L_n$&$ U_n$&$ U_n\circ
L_n$ &$L_n\circ U_n$\\\hline $ U_n\circ L_n$&$ U_n\circ
L_n$&$L_n\circ U_n$&$U_n\circ L_n$&$L_n\circ U_n$\\\hline $
L_n\circ U_n$&$U_n\circ L_n$&$ L_n\circ U_n$ &$U_n\circ L_n$ &$
L_n\circ U_n$\\\hline
\end{tabular}
\end{center}
\caption{LULU semi-group}
\end{table}

Let us recall that, according to the well known theorem of
Matheron \cite{Matheron}, in general, two ordered morphological
operators generate a six element semi-group which is only
partially ordered.

The power of the LULU operators as separators is further
demonstrated by their Total Variation Preservation property. Let
$BV(\mathbb{Z})$ be the set of sequences with bounded variation,
that is,%
\[
BV(\mathbb{Z})=\{\xi\in\mathcal{A}(\mathbb{Z}):\sum_{i\in\mathbb{Z}}|\xi_i-\xi_{i+1}|<\infty\}.
\]
Total Variation of a sequence $\xi\in BV(\mathbb{Z})$ is given by
$TV(\xi)=\sum\limits_{i\in\mathbb{Z}}|\xi_i-\xi_{i+1}|$.

\begin{definition}\label{defTVP}
An operator $P:BV(\mathbb{Z})\to BV(\mathbb{Z})$ is called {\bf
total variation preserving} if
\begin{equation}\label{defTVP1}
TV(\xi)=TV(P(\xi))+TV((id-P)(\xi))\ ,\ \ \xi\in BV(\mathbb{Z}).
\end{equation}
\end{definition}
We should note that since $TV$ is a semi-norm on $BV(\mathbb{Z})$
we always have
\[
TV(\xi)\leq TV(P(\xi))+TV((id-P)(\xi)).
\]
Hence, the significance of the equality (\ref{defTVP1}) is  that
the decomposition $f=A(f)+(id-A)(f)$ does not create additional
total variation. In particular, this property is very important
for the application of the LULU operators to discrete pulse
decompositions of sequences.

The aim of this paper is to generalize the LULU operators to
functions on $\mathbb{Z}^d$ in such a way that their essential
properties are preserved. In Section 2 the definitions of the
basic operators $L_n$ and $U_n$ on $\mathcal{A}(\mathbb{Z}^d)$ are
derived by using a strengthened form of the morphological concept
of connection. Then we show that indeed these operators replicate
the properties of the LULU operators for sequence. More precisely,
we prove that: (i) they are separators (Section 2); (ii) their
smoothing effect can be described in a similar way to the
$n$-monotonicity of sequences (Section 3); (iii) they generate a
four element fully ordered semi-group (Section 4). The developed
theory can be applied to many problems of Image Analysis and it is
the intention of the authors to research such applications in the
future. However, as an illustration and demonstration of the power
of this approach we apply the newly defined operators to deriving
a total variation preserving discrete pulse decomposition of
images. Noise removal and partial reconstructions are discussed in
Section 6.

\section{The basic operators $L_n$ and $U_n$.}\label{definitions}

The definition of the operators $L_n$ and $U_n$ for sequences
involves maxima and minima over sets of consecutive terms, thus,
making an essential use of the fact that $\mathbb{Z}$ is totally
ordered. Since $\mathbb{Z}^d$, $d>1$, is only partially ordered
the concept of `consecutive' does not make sense in this setting.
Instead, we use the morphological concept of set connection,
\cite{Serra88}.
\begin{definition}\label{defConection}
Let $B$ be an arbitrary non-empty set. A family $\mathcal{C}$ of
subsets of $B$ is called a connected class or a {\bf connection}
on $B$
if\\
(i) $\emptyset\in\mathcal{C}$\\
(ii) $\{x\}\in\mathcal{C}$ for all $x\in B$\\
(iii) for any family $\{C_i:i\in\mathcal{C}\}\subseteq
\mathcal{C}$
\[
\bigcap_{i\in I}C_i\neq\emptyset \Longrightarrow \bigcup_{i\in
I}C_i\in\mathcal{C}
\]
\end{definition}
This definition generalizes the topological concept of
connectivity to arbitrary sets including discrete sets like
$\mathbb{Z}^d$. If a set $C$ belongs to a connection $\mathcal{C}$
then $C$ is called \emph{connected}.

It is clear from Definition \ref{defConection} that a connection
on $\mathbb{Z}^d$ does not necessarily contain sets of every size.
For example, \mbox{$\{ \emptyset\}\cup\{\{x\}:x\in
\mathbb{Z}^d\}$} and
\linebreak\mbox{$\{\emptyset\}\cup\{\{x\}:x\in
\mathbb{Z}^d\}\cup\{\mathcal{Z}^d\}$} are connections on
$\mathbb{Z}^d$ but neither of them contain sets of finite size
other than 0 and 1. In the definition of the operators $L_n$ and
$U_n$ we need sets of every size. We assume that the set
$\mathbb{Z}^d$ is equipped with a connection $\mathcal{C}$ which
satisfies the following three conditions
\begin{eqnarray}
&\bullet&\mathbb{Z}^d\in\mathcal{C}\label{CondCon1}\\
&\bullet&\mbox{For any }a\in\mathbb{Z}^d,
E_a(C)\in\mathcal{C}\mbox{ whenever }C\in\mathcal{C}\label{CondCon2}\\
&\bullet&\mbox{If } V\subsetneq W,\ V,W\in\mathcal{C}, \mbox{ then
there exists }x\in W\setminus V\nonumber\\&& \mbox{ such that
}V\cup \{x\}\in\mathcal{C}\label{CondCon3}
\end{eqnarray}
The aim of the conditions (\ref{CondCon1})--(\ref{CondCon3}) is to
define a connection which is sufficiently rich in connected sets.
This is demonstrated by the following property, which is obtained
via iterative application of the property (\ref{CondCon3}):
\begin{eqnarray}
&\bullet&\mbox{Let }V\subsetneq W,\ V,W\in\mathcal{C}.\mbox{ For
every }k\in\mathbb{N} \mbox{ such that } \nonumber\\&&
\card(V)<k<\card(W)\mbox{ there exists }S\in \mathcal{C}\\&&\mbox{
such that }V\subseteq S\subseteq W \mbox{ and
}\card(S)=k.\nonumber \label{CondCon4}
\end{eqnarray}

As usual, $\card(V)$ is the number of the elements in the set $V$,
that is, the size of $V$. For $V\subseteq\mathbb{Z}^d$ we have
$\card(V)\in\mathbb{N}\cup\{0\}\cup\{\infty\}$. Given a point
$x\in \mathbb{Z}^d$ and $n\in\mathbb{N}$ we denote by
$\mathcal{N}_n(x)$ the set of all connected sets of size $n+1$,
which contain point $x$, that is,
\begin{equation}\label{Nn}
\mathcal{N}_n(x)=\{V\in\mathcal{C}:x\in V,\ \card(V)=n+1\}.
\end{equation}
Now the operators $L_n$ and $U_n$ are defined on
$\mathcal{A}(\mathbb{Z}^d)$ as follows.

\begin{definition}\label{defLnUn}
Let $f\in\mathcal{A}(\mathbb{Z}^d)$ and $n\in\mathbb{N}$. Then
\begin{eqnarray}
L_n(f)(x)&=&\max_{V\in\mathcal{N}_n(x)}\min_{y\in V}f(y),\ x\in\mathbb{Z}^d,\label{defLULU1}\\
U_n(f)(x)&=&\min_{V\in\mathcal{N}_n(x)}\max_{y\in V}f(y),\
x\in\mathbb{Z}^d.\label{defLULU2}
\end{eqnarray}
\end{definition}

Let us first see that Definition \ref{defLnUn} generalizes the
definition of $L_n$ and $U_n$ for sequences. Suppose $d=1$ and let
$\mathcal{C}$ be the connection on $\mathbb{Z}$ generated by the
pairs of consecutive numbers. Then all connected sets on
$\mathbb{Z}$ are sequences of consecutive integers and for any
$i\in\mathbb{Z}$ we have
\[
\mathcal{N}_n(i)=\{\{i\!-\!n,i\!-\!n\!+\!1,...,i\},\{i\!-\!n\!+\!1,i\!-\!n\!+\!2,...,i\!+\!1\},...,\{i,i\!+\!1,...,i\!+\!n\}\}
\]
Hence for an arbitrary sequence $\xi$ considered as a function on
$\mathbb{Z}$ the formulas (\ref{defLULU1}) and (\ref{defLULU2})
are reduced to (\ref{LnSec}) and (\ref{UnSec}), respectively.

\begin{theorem}\label{theoOrderProp}
{\bf (Order Properties)}
\begin{itemize}
\item[a)] $L_n\leq id\leq U_n$

\item[b)] $f\leq g\Longrightarrow (\ L_n(f)\leq L_n(g),\
U_n(f)\leq U_n(g)\ )$

\item[c)] $n_1<n_2 \Longrightarrow (\ L_{n_1}\geq L_{n_2},\
U_{n_1}\leq U_{n_2}\ )$
\end{itemize}
\end{theorem}
\begin{proof} We will only prove the inequalities involving $L_n$
since those involving $U_n$ are proved similarly.

\noindent a) Let $f\in\mathcal{A}(\mathbb{Z}^d)$. For every $x \in
Z^d$ and $V \in \mathcal{N}_n(x)$ we have
\[\min_{y \in V}f(y) \le f(x). \]
Hence
\[
L_n(f)(x)=\max_{V\in \mathcal{N}_n(x)}\min_{y \in V}f(y) \le
f(x),\ x\in\mathbb{Z}^d.
\]
Therefore, $L_n(f)\leq f$, $f\in\mathcal{A}(\mathbb{Z}^d)$, which
implies $L_n\leq id$.

\noindent b) Let $f \le g$. For any $x \in Z^d$ and $V \in
\mathcal{N}_n(x)$, we have $\min\limits_{y \in V}f(y) \le
\min\limits_{y \in V}
g(y)$. Therefore %
\[
L_n(f)(x)=\max_{V \in \mathcal{N}_n(p)} \min_{y \in V}f(y) \le
\max_{V \in \mathcal{N}_n(p)} \min_{y \in V}g(y)=L_n(g)(x),\ x \in
Z^d.
\]

\noindent c) Let $f\in\mathcal{A}(\mathbb{Z}^d)$. It follows from
(\ref{CondCon4}) that for every $x \in Z^d$ and $V \in
\mathcal{N}_{n_2}(x)$ there exists a set $W \in
\mathcal{N}_{n_1}(x)$ such that $W \subseteq V$. Therefore
\[
\min_{y \in V}f(y) \le \min_{y\in W}f(y) \le \max_{S \in
\mathcal{N}_{n_1}(x)} \min_{y \in S} f(y) = L_{n_1}(f)(x).
\]
Hence
\[
L_{n_2}(f)(x) = \max_{V \in \mathcal{N}_{n_2}(x)} \min_{y \in V}
f(y) \le L_{n_1}(f)(x), x\in\mathbb{Z}^d.
\]
\end{proof}

\begin{theorem}\label{theoSep}For any $n\in\mathbb{N}$ the
operators $L_n$ and $U_n$ are separators.
\end{theorem}
\begin{proof}
We will only verify the conditions (i)--(v) in Definition
\ref{defSep} for $L_n$ since $U_n$ is dealt with in a similar
manner.

\noindent(i) Let $a\in \mathbb{Z}^d$ and
$f\in\mathcal{A}(\mathbb{Z}^d)$. Using the property
(\ref{CondCon2}), for every $x\in\mathbb{Z}^d$ we have
\[
\mathcal{N}_n(x+a)=a+\mathcal{N}_n(x)=\{a+V:V\in\mathcal{N}_n(x)\}
\]
Therefore,%
\begin{eqnarray*}
E_a(L_n(f))(x)&=&L_n(f)(x+a)=\max_{V \in \mathcal{N}_n(x+a)}
\min_{y \in V} f(y) \\
&=&\max_{V \in \mathcal{N}_n(x)} \min_{y \in a+V} f(y)
\ =\ \max_{V \in \mathcal{N}_n(x)} \min_{y \in V} f(y+a) \\
&=&\max_{V \in \mathcal{N}_n(x)} \min_{y \in V} E_a(f)(y),\
x\in\mathbb{Z}^d
\end{eqnarray*}

\noindent (ii) Let  $f,c\in\mathcal{A}(\mathbb{Z}^d)$, where $c$
is a constant function with a value of $\theta$. Then for every
$x\in\mathbb{Z}^d$ we have
\begin{eqnarray*}
L_n(f+c)(x) &=& \max_{V \in \mathcal{N}_n(x)}\min_{y \in V}
(f+c)(y) \ =\ \max_{V \in
\mathcal{N}_n(x)}\min_{y \in V} (f(y)+\theta) \\
&=&\left(\max_{V \in \mathcal{N}_n(x)}\min_{y \in V}
f(y)\right)+\theta \ =\ L_n(f)(x) +c(x)
\end{eqnarray*}

\noindent (iii) Let  $f\in\mathcal{A}(\mathbb{Z}^d)$ and
$\alpha\in\mathbb{R}$, $\alpha\geq 0$. For every
$x\in\mathbb{Z}^d$ we have
\[
L_n(\alpha f)(x) = \max_{V \in \mathcal{N}_n(x)}\min_{q \in V}
(\alpha f )(y) = \alpha\left( \max_{V \in \mathcal{N}_n(x)}\min_{q
\in V} f(y)\right) = \alpha L_n(f)(x).
\]

\noindent (iv) The inequality \[L_n \circ L_n \le L_n\] is an
immediate consequence of Theorem \ref{theoOrderProp}. Then it is
sufficient to prove the inverse inequality. Let $f \in
\mathcal{A}(\mathbb{Z}^d)$ and $x \in \mathbb{Z}^d$. We have
\begin{equation}\label{theoSep1}
L_n(L_n(f))(x) = \max_{W \in \mathcal{N}_n(x)} \min_{y \in W}
\max_{V \in \mathcal{N}_n(y)} \min_{z \in V} f(z).
\end{equation}
But $y \in W\in \mathcal{N}_n(x)$ implies $W \in
\mathcal{N}_n(y)$. Therefore for every $W\in \mathcal{N}_n(x)$ and
$y\in W$ we have
\[
\max_{V \in \mathcal{N}_n(y)} \min_{z \in V}f(z) \ge \min_{z \in
W} f(z).
\]
Using that the right hand side is independent of $y$ we further
obtain
\[
\min_{y \in W} \max_{V \in \mathcal{N}_n(y)} \min_{z \in V} \ge
\min_{z \in W} f(z),\ W\in \mathcal{N}_n(x).
\]
Then it follows from the representation (\ref{theoSep1}) that
\[
L_n(L_n(f))(x) \ge \max_{W \in \mathcal{N}_n(x)} \min_{z \in W}
f(z) = L_n(f)(x).
\]

\noindent (v) The co-idempotence of the operator $L_n$ is
equivalent to $L_n\circ(id-L_n)=0$. The inequality
$L_n\circ(id-L_n)\geq 0$ is an easy consequence of Theorem
\ref{theoOrderProp}. Hence, for the co-idempotence of $L_n$ it
remains to show that $L_n\circ(id-L_n)\leq 0$. Assume the
opposite. Namely, there exists a function
$f\in\mathcal{A}(\mathbb{Z}^d)$ and $x\in\mathbb{Z}^d$ such that
$(L_n\circ(id-L_n))(f)(x)>0$. Using the definition of $L_n$ this
inequality implies that there exists $V\in \mathcal{N}_n(x)$ such
that for every $y\in V$ we have $(id-L_n)(f)(z)>0$,
or equivalently %
\begin{equation}\label{theoSep2}
f(y)>L_n(f)(y),\ y\in V.
\end{equation}
Let $z\in V$ be such that $f(z)=\min\limits_{t\in V} f(t)$. Then
for every $y\in V$ we have
\begin{equation}\label{theoSep3}
L_n(f)(y)=\max_{W\in\mathcal{N}_n(y)}\min_{t\in W}f(t)\geq
\min_{t\in V}f(t)\geq f(z).
\end{equation}
Taking $y=z$ in (\ref{theoSep2}) and (\ref{theoSep3}) we obtain a
contradiction which completes the proof.
\end{proof}

\section{The operators $L_n$ and $U_n$ as smoothers}

Similar to their counterparts for sequences the operators $L_n$
and $U_n$ defined in Section \ref{definitions} smooth the input
function by removing sharp peaks (the application of $L_n$) and
deep pits (the application of $U_n$). The smoothing effect of
these operations is made more precise by using the concepts of a
local maximum set and a local minimum set given below.

\begin{definition}\label{defAdj}
Let $V\in\mathcal{C}$. A point $x$ is called {\bf adjacent} to $V$
if $V\cup\{x\}\in\mathcal{C}$. The set of all points adjacent to
$V$ is denoted by $adj(V)$, that is,
\[
\adj(V)=\{x\in\mathbb{Z}^d:V\cup\{x\}\in\mathcal{C}\}.
\]
\end{definition}

An equivalent formulation of the property (\ref{CondCon3}) of the
connection $\mathcal{C}$ is as follows:
\begin{equation}\label{AdjProp}
V,W\in\mathcal{C},\ W\subsetneq V\ \Longrightarrow\  \adj(W)\cap
V\neq \emptyset.
\end{equation}

\begin{definition} \label{defLocMaxMin}
A connected subset $V$ of $\mathbb{Z}^d$ is called a {\bf local
maximum set} of $f \in \mathcal{A}(Z^d)$ if
\[
\max_{y \in \adj(V)}f(y) < \min_{x \in V} f(x).
\]
Similarly $V$ is a {\bf local minimum set} if
\[
\min_{y \in\adj(V)}f(y) > \max_{x \in V} f(x).
\]
\end{definition}

The next four theorems deal with different aspects of the
application of $L_n$ and $U_n$ to functions in
$\mathcal{A}(\mathbb{Z}^d)$. Their cumulative effect will be
discussed at the end of the section. All theorems contain
statements a) and b). Due to the similarity we present only the
proofs of a).

\begin{theorem} \label{theoLocMaxMin1} Let
$f\in\mathcal{A}(\mathbb{Z}^d)$ and $x\in\mathbb{Z}^d$. Then we have\\
a) $L_n(f)(x) < f(x)$ if and only if there exists a local maximum
set $V$ such that $x\in V$ and $\card(V)\leq n$;\\
b) $U_n(f)(x) > f(x)$ if and only if there exists local minimum
set $V$ such that $x\in V$ and $\card(V)\leq n$.
\end{theorem}
\begin{proof}a)
\underline{Implication to the left.} Suppose that there exists a
local maximum set $V \in \mathcal{N}_k(x)$, $k<n$. Consider an
arbitrary $W \in \mathcal{N}_n(x)$ and let $S=W \cap V$. Then,
since the size of $W$ is larger than the size of $S$ we have $W
\setminus S \ne \emptyset$. Furthermore, by (\ref{AdjProp}) we
have $\adj(S) \cap W \ne \emptyset$. Let $z \in \adj(S) \cap W$.
Since $\adj(S) \cap W \subseteq W\setminus S = W \setminus V$, we
have that $z \notin V$ but $z \in \adj(V).$ Then using also that
$V$ is a local maximum set we obtain
\[
\min_{y \in W}f(y) \le f(z) < \min_{t \in V}f(t) \le f(x).
\]
Since the set $W \in \mathcal{N}_n(p)$ is arbitrary, this
inequality implies that $L_n(f)(x) < f(x)$.

\noindent\underline{Implication to the right.} Suppose $L_n(f)(x)
< f(x)$. Let $V$ be the largest (in terms of $\subseteq$)
connected set containing $x$ such that
\begin{equation}\label{theoLMM01}
f(y) \ge f(x),\ y \in V.
\end{equation}
The set $V$ is obviously unique and can be constructed as
$V=\gamma_x(Y)$, where $\gamma_x$ is the morphological point
connected opening generated by $x$, see \cite{Serra88} or
\cite{Serra06}, and $Y = \{y \in \mathbb{Z}^d: f(y) \ge f(x)\}$.
We have $f(z)<f(x)$, $z\in \adj(V)$, because otherwise
(\ref{theoLMM01}) is satisfied on the larger connected set
$\{z\}\cap V$. Therefore
\[
\max_{z\in\adj(V)}f(z)<f(x)=\min_{y\in V}f(y).
\]
Hence $V$ is a local maximum set.

Assume that $\card(V) > n$. It follows from (\ref{CondCon4}) that
there exists $W \in \mathcal{N}_n(x)$ such that $W \subset V$.
Then
\[
L_n(f)(x)=\max_{V \in \mathcal{N}_n(x)}\min_{y \in V} f(y) \ge
\min_{y\in W}f(y)\ge \min_{y \in V}f(y) = f(x).
\]
This contradicts the assumption $L_n(f)(x)<f(x)$. Therefore,
$\card(V)\leq n$.
\end{proof}

\begin{theorem} \label{theoLUsmooth}
Let $f \in \mathcal{A}(\mathcal{Z}^d)$. Then\\
a) the size of any local maximum set of the function $L_n(f)$ is
larger than $n$;\\
b) the size of any local minimum set of the function $U_n(f)$ is
larger than $n$.
\end{theorem}
\begin{proof}a) Assume the opposite, that is,
there exists a local maximum set $V$ of $L_n(f)$ such that
$\card(U)\leq n$. By Theorem \ref{theoLocMaxMin1} we have that
\[
L_n(L_n(f))(x) < L_n(f)(x),\ x\in V.
\]
Since $L_n$ is idempotent, see Theorem \ref{theoSep}, this implies
the impossible inequality  $L_n(f)(x) < L_n(f)(x)$, which
completes the proof.
\end{proof}

\begin{theorem}\label{theoMon}
Let $V\in\mathcal{C}$ and let $x\in\adj(V)$.\\[3pt]
a) If $f(x)\leq \min\limits_{y\in V}f(y)$ then $L_n(f)(x)\leq
\min\limits_{y\in V}L_n(f)(y)$;\\[3pt]
b) If $f(x)\geq \max\limits_{y\in V}f(y)$ then $U_n(f)(x)\geq
\max\limits_{y\in V}U_n(f)(y)$.
\end{theorem}

\begin{proof}a) For any $W\in\mathcal{N}_n(x)$ the set $W\cup V$ is
connected and of size larger than $n+1$. Therefore, by
(\ref{CondCon4}), for every $y\in V$ there exists
$S_y\in\mathcal{N}_n(y)$ such that $S_y\subset W\cup V$. Then,
using also the given inequality, for every $y\in V$ and
$W\in\mathcal{N}_n(q)$ we have
\[
\min_{z \in W} f(z)=\min_{z \in W\cup V} f(z)\leq
 \min_{z\in S_y} f(z)\leq L_n(f)(y).
 \]
Hence
\[
L_n(f)(x) = \max_{W \in \mathcal{N}_n(x)}\min_{z \in W} f(z)\leq
\min_{y\in V}L_n(f)(y).
\]
\end{proof}

\begin{theorem} \label{theoNotNewLM}
Let $f \in \mathcal{A}(\mathcal{Z}^d)$ and $V \in\mathcal{C}$.\\
a) If $V$ is a local minimum set of $L_n(f)$ then there exists a
local minimum set $W$ of $f$ such that $W \subseteq V$.\\
b) If $V$ is a local maximum set of $U_n(f)$ then there exists a
local maximum set $W$ of $f$ such that $W \subseteq V$.
\end{theorem}
\begin{proof} a)
Let $V$ be a local minimum set of $L_n(f)$. Then
\[
\min_{y\in\adj(V)}f(y) \ge \min_{y\in\adj(V)}L_n(f)(y) > L_n(f)(x)
~\forall~x\in V.
\]
Let $q\in\adj(V)$ be such that
$f(q)=\min\limits_{y\in\adj(V)}f(y)$ and let \[Y=\{y \in V: f(y)
<f(q)\}.\] An easy application of Theorem \ref{theoMon} shows that
$Y\neq\emptyset$. Let $t\in Y$ and let $W$ be the largest (with
respect to inclusion) connected subset of $Y$ which contains $t$.
As in the proof of Theorem \ref{theoLocMaxMin1}, the set $W$ can
be obtained through $W=\gamma_t(Y)$. For every $z\in\adj(W)$ we
have $f(z)\geq f(q)>\max_{y\in W}f(y)$. Therefore $W$ is a local
minimum set of $f$.
\end{proof}

Theorems \ref{theoLocMaxMin1}--\ref{theoNotNewLM} provide the
following characterization of the effect of the operators $L_n$
and $U_n$ of a function
$f\in\mathcal{A}(\mathbb{Z}^d)$:

\begin{itemize}

\item The application of $L_n$ ($U_n$) removes local maximum
(minimum) sets of size smaller or equal to $n$.

\item The operator $L_n$ ($U_n$) does not affect the local minimum
(maximum) sets in the sense that such sets may be affected only as
a result of the removal of local maximum (minimum) sets. However,
no new local minimum sets are created where there were none. This
does not exclude the possibility that the action of $L_n$ ($U_n$)
may enlarge existing local maximum (minimum) sets or join two or
more local maximum (minimum) sets of $f$ into one local maximum
(minimum) set of $L_n(f)$ ($U_n(f)$).

\item $L_n(f)=f$ ($U_n(f)=f$) if and only if $f$ does not have
local maximum (minimum) sets of size $n$ or less;

\end{itemize}

Furthermore, as an immediate consequence of Theorem
\ref{theoLUsmooth} and Theorem \ref{theoNotNewLM} we obtain the
following corollary.

\begin{corollary}\label{theoLUcompsmooth}
For every $f\in\mathcal{A}(\mathbb{Z}^d)$ the functions $(L_n\circ
U_n)(f)$ and $(U_n\circ L_n)(f)$ have neither local maximum sets
nor local minimum sets of size $n$ or less. Furthermore,
\[
(L_n\circ U_n)(f)=(U_n\circ L_n)(f)=f
\]
if and only if $f$ does not have local maximum sets or local
minimum sets of size less than or equal to $n$.
\end{corollary}

We should remark that in the one dimensional setting, the
sequences without local maximum sets or local minimum sets of size
less than or equal to $n$ are exactly the so-called $n$-monotone
sequences. Hence Corollary \ref{theoLUcompsmooth} generalizes the
respective results in the LULU theory of sequences, \cite[Theorem
3.3]{Rohwerbook}.

\section{The LULU semi-group}

In this section we consider the operators $L_n$, $U_n$ and their
compositions. The main result is that $L_n$, $U_n$, $L_n\circ U_n$
and $U_n\circ L_n$ form a semi-group with respect to composition
with a composition table as given in Table 1. Furthermore, the
semi-group is totaly ordered as in (\ref{LULUorder}) with respect
to the point-wise defined partial order (\ref{defOrder}).

\begin{theorem}\label{theoLUidemp}
The operators $L_n\circ U_n$ and $U_n\circ L_n$ are idempotent,
that is,
\begin{eqnarray}
&&L_n\circ U_n\circ L_n\circ U_n=L_n\circ U_n\ ,\label{theoLUidemp1}\\
&&U_n\circ L_n\circ U_n\circ L_n=U_n\circ L_n\
.\label{theoLUidemp2}
\end{eqnarray}
\end{theorem}
\begin{proof}
Using the order properties in Theorem \ref{theoOrderProp} and the
idempotence of $L_n$ and $U_n$, see Theorem \ref{theoSep}, we have
\begin{eqnarray*}
&&L_n\circ U_n\circ L_n\circ U_n\leq L_n\circ U_n\circ id\circ
U_n=L_n\circ U_n\circ U_n=L_n\circ U_n\\
&&L_n\circ U_n\circ L_n\circ U_n\geq L_n\circ id\circ L_n\circ
U_n=L_n\circ L_n\circ U_n=L_n\circ U_n
\end{eqnarray*}
which implies (\ref{theoLUidemp1}). The equality
(\ref{theoLUidemp2}) is proved similarly.
\end{proof}

\begin{theorem}\label{theoGroupMain}For any $n\in\mathbb{N}$ we
have
\begin{equation}\label{theoGroupMain01}
L_n\circ U_n \circ L_n = U_n \circ L_n.
\end{equation}
\end{theorem}
\begin{proof}  It follows from Theorem \ref{theoOrderProp} that
\begin{equation}\label{theoGroupMain02}
L_n\circ U_n \circ L_n \le id\circ U_n \circ L_n=U_n \circ L_n.
\end{equation}
Assume that (\ref{theoGroupMain01}) is violated. In view of
(\ref{theoGroupMain02}), this means that there exists $f \in
\mathcal{A}(\mathbb{Z}^d)$ and $z \in \mathbb{Z}^d$ such that
\[
L_n (U_n(L_n(f)))(z) < U_n(L_n(f))(z).
\]
It follows from Theorem \ref{theoLocMaxMin1} that there exists
$k\leq n$ and $V \in \mathcal{N}_k(z)$ such that $V$ is a local
maximum set for $U_n(L_n(f))(z)$. Then, by Theorem
\ref{theoNotNewLM}, there exists $W\subseteq V$ such that $W$ is a
local maximum set of the function $L_n(f)$. We have $\card(W)\leq
k\leq n$. However, $L_n(f)$ does not have any local maximum sets
of size less than or equal to $n$, see Theorem \ref{theoLUsmooth}.
This contradiction completes the proof.
\end{proof}

As in the case of sequences, the key result for the set
\begin{equation}\label{group}
\{L_n,\ U_n,\ L_n\circ U_n,\ U_n\circ L_n\}
\end{equation}
to be closed under composition is the equality in Theorem
\ref{theoGroupMain}. Now one can easily derive the rest of the
formulas for the compositions of the operators in this set. The
composition table is indeed as given in Table 1. Furthermore,
Theorem \ref{theoGroupMain} implies the total order on the set
(\ref{group}) as in (\ref{LULUorder}). Indeed, we have
\[
L_n=id\!\circ\! L_n\leq U_n\!\circ\! L_n=L_n\!\circ\! U_n\!\circ\!
L_n\leq L_n\!\circ\! U_n\!\circ id=L_n\!\circ\! U_n\leq
id\!\circ\! U_n=U_n
\]
Therefore, the operators $L_n$ and $U_n$ for functions on
$\mathbb{Z}^d$ generate via composition a semi-group with exactly
the same algebraic and order structure as the semi-group generated
by the operators $L_n$ and $U_n$ for sequences.

\section{Discrete pulse transform of images}

In this section we apply the LULU operators defined and
investigated in the preceding sections to derive a discrete pulse
decomposition of images. A grayscale image is given through a
function $f$ on a rectangular domain $\Omega\subset\mathbb{Z}^2$,
the value of $f$ being the luminosity at the respective pixel. For
the theoretical study it is more convenient to assume that the
functions are defined on the whole space $\mathbb{Z}^2$. To this
end one can, for example, define $f$ on the set
$\mathbb{Z}^2\setminus\Omega$ as a constant, e.g. 0. Hence we
consider the set $\mathcal{A}(\mathbb{Z}^2)$.

Appropriate connections for images are defined through a relation
$r$ on $\mathbb{Z}^2$ reflecting what we consider neighbors of a
pixel in the given context. Figure 1 gives some examples of the
the neighbors of the pixel $(i,j)$.

\begin{figure}[h]
\includegraphics[scale=0.2, width=4.5cm, height=4.5cm]{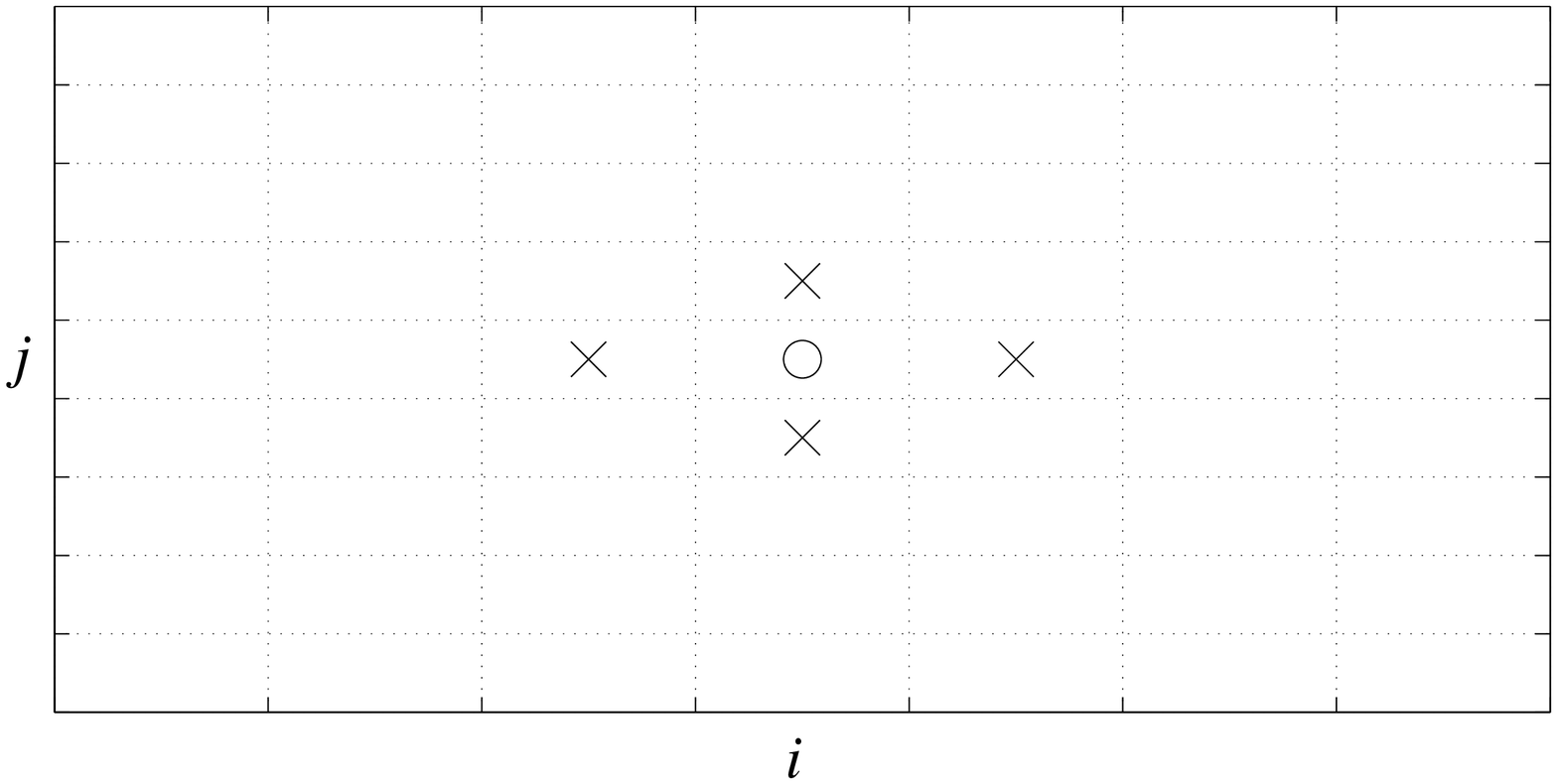}
\includegraphics[scale=0.2, width=4.5cm, height=4.5cm]{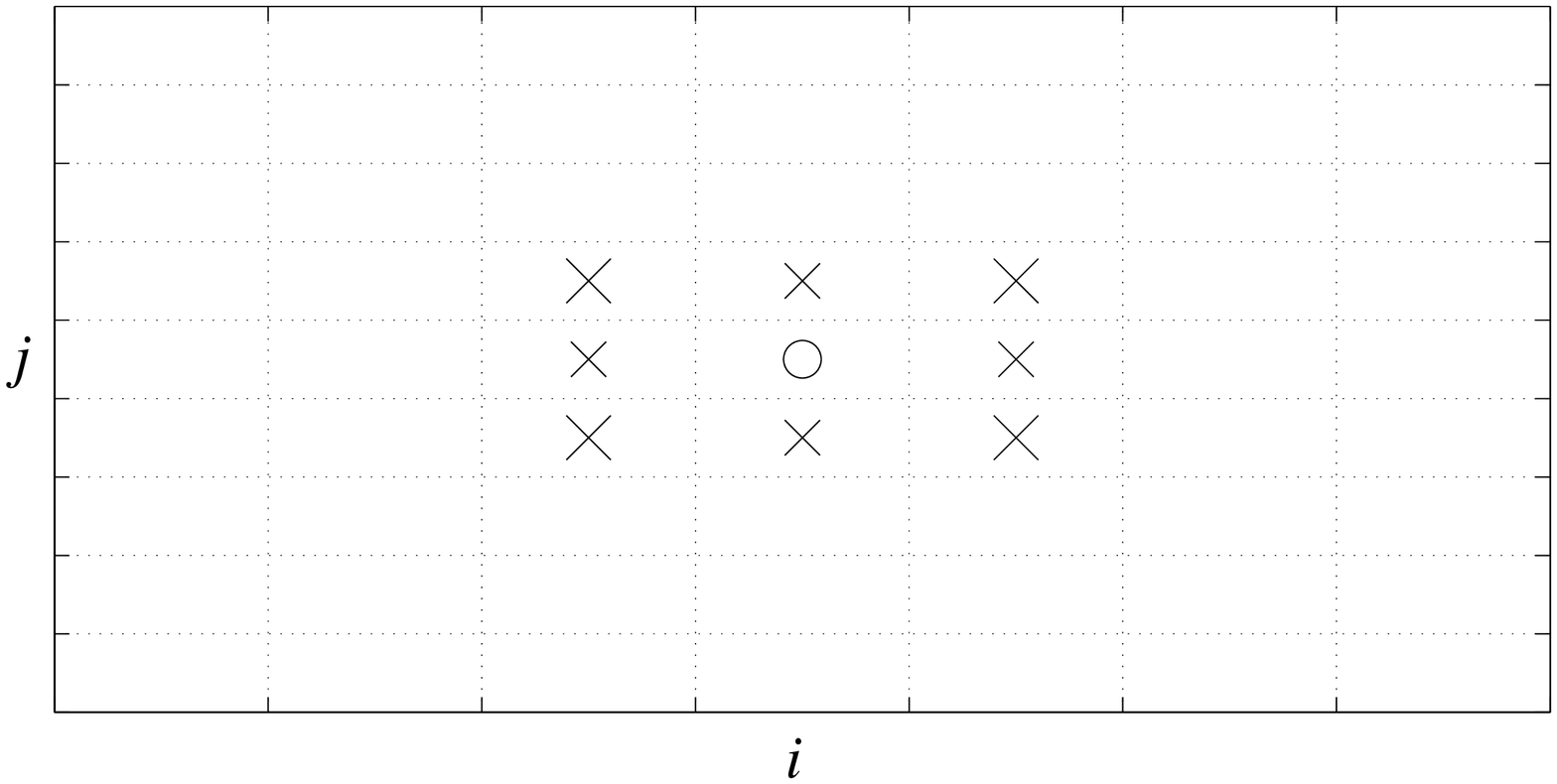}
\includegraphics[scale=0.2, width=4.5cm, height=4.5cm]{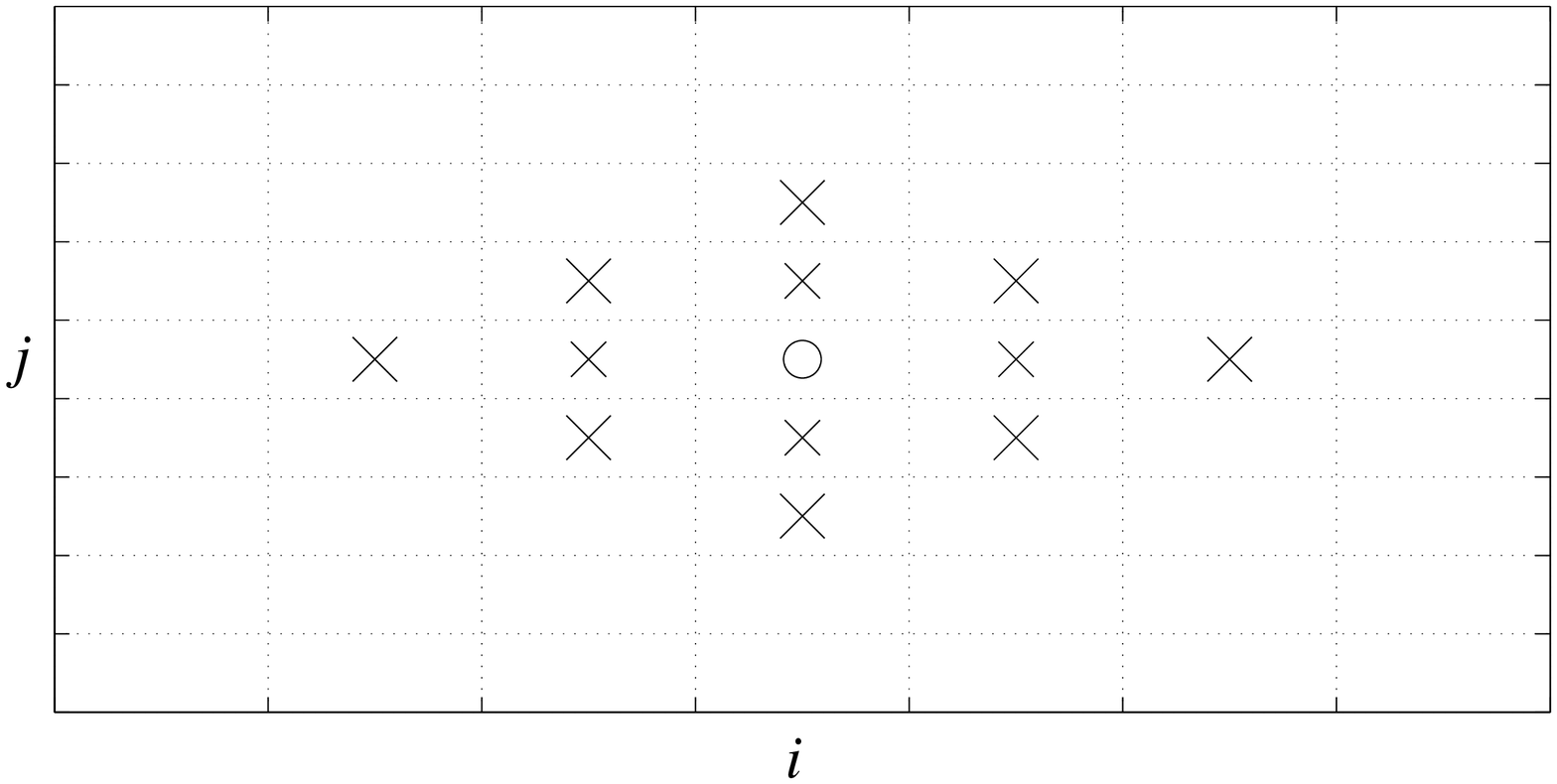}
\caption{Neighbors of $(i,j)$}
\end{figure}

We call a set $C\subseteq\mathbb{R}^2$ \emph{connected} if for any
two pixels $p,q\in C$ there exists a set of pixels
$\{p_1,p_2,...,p_k\}\subseteq C$ such that each pixel is neighbor
to the next one, $p$ is neighbor to $p_1$ and $p_k$ is neighbor to
$q$. We assume that the neighbor relation $r$ on $\mathbb{Z}^2$ is
such that
\begin{eqnarray}
&\bullet&r \mbox{ is reflexive, symmetric and shift
invariant}\label{NeighborCond1}\\
&\bullet&((i,j),(i\pm 1,j))\in r \mbox{ and }((i,j),(i,j\pm 1))\in
r, \mbox{ for all }i,j\in\mathbb{Z}.\label{NeighborCond2}
\end{eqnarray}
The conditions (\ref{NeighborCond1})--(\ref{NeighborCond2}) ensure
that the set of connected set $\mathcal{C}$ defined through this
relation is a connection in terms of Definition \ref{defConection}
and satisfies the conditions (\ref{CondCon1})--(\ref{CondCon3}).
Hence we can apply the operators $L_n$ and $U_n$ discussed in the
preceding sections to functions on $\mathbb{Z}^2$. Similar to the
case of sequences we obtain a decomposition of a function
$f\in\mathcal{A}(\mathbb{Z}^2)$ by applying iteratively the
operators $L_n,U_n$ with $n$ increasing from 1 to $\infty$. This
can be done in different ways depending on sequencing of the $L_n$
and $U_n$. Since this section is intended as a demonstration
rather than presenting a comprehensive discrete pulse transform
theory, we will take one particular case when $U_n$ follows $L_n$.
Define the operators $F_n$, $n\in\mathbb{N}$, by $F_1=U_1\circ
L_1$ and $F_n=U_n\circ L_n\circ F_{n-1}$. Then for any
$f\in\mathcal{A}(\mathbb{Z}^2)$ and $m\geq 1$ we have
\begin{eqnarray}
&&\hspace{-1.5cm}f=(id-U_1\!\circ\! L_1)(f)+((id-U_2\!\circ\!
L_2)\!\circ\!
F_1)(f)+((id-U_3\!\circ\! L_3)\!\circ\! F_2)(f)\nonumber\\
&&\ \ \ +\ ...\ +((id-U_m\!\circ\! L_m)\!\circ\!
F_{m-1})(f)+F_m(f)\label{decomp}
\end{eqnarray}

\begin{definition}\label{defDPulse}
A function $\phi\in\mathcal{A}(\mathbb{Z}^2)$ is called a pulse if
there exist a connected set $V$ and a real number $\alpha$ such
that
\[
\phi(x)=\left\{\begin{tabular}{ccc} $\alpha$&if&$x\in
V$\\0&if&$x\in\mathbb{Z}^2\setminus V$.\end{tabular}\right.
\]
The set $V$ is called support of the pulse $\phi$ and is denoted
by $\supp(\phi)$.
\end{definition}

Figure 2 gives an example of a pulse. It should be remarked that
the support of a pulse may generally have any shape, the only
restriction being that it is connected.

\begin{figure}[h]
\begin{center}
\includegraphics[scale=0.4]{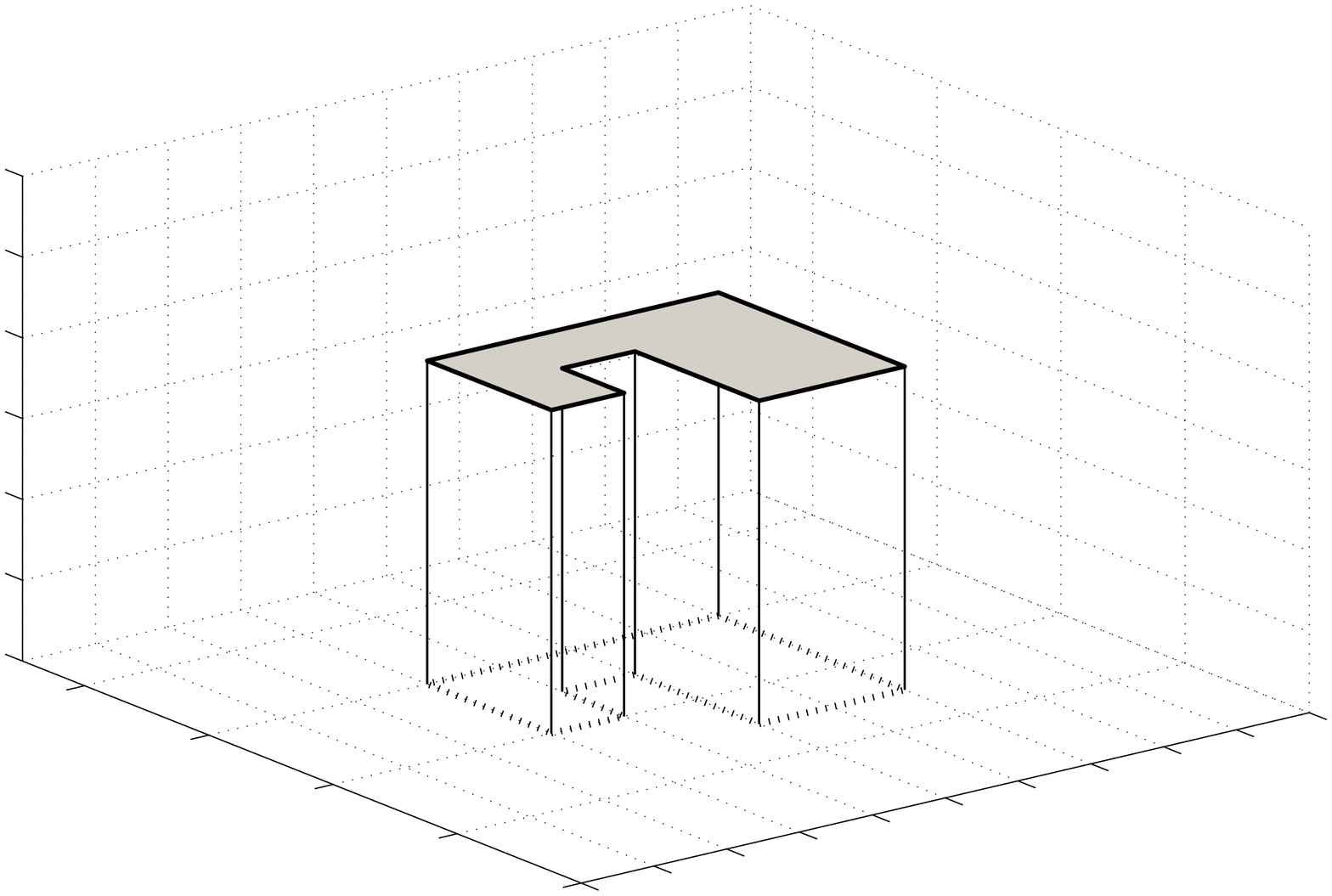}
\caption{Pulse}
\end{center}
\end{figure}

The usefulness of the representation (\ref{decomp}) of a function
$f\in\mathcal{A}(\mathbb{Z}^2)$ is in the fact that all terms are
sums of pulses as stated in the next theorem.

\begin{theorem}\label{theoDPT}
Let $f\in\mathcal{A}(\mathbb{Z}^2)$.
\begin{itemize}\item[a)] For every $n\in\mathbb{N}$ the function $((id-U_n\circ L_n)\circ
F_{n-1}(f)$ is a sum of discrete pulses with disjoint support,
that is, there exist $\gamma(n)\in\mathbb{N}$ and discrete pulses
$\phi_{ns}$, $s=1,...,\gamma(n)$, such that
\begin{equation}\label{theoDPTprop0}
((id-U_n\circ L_n)\circ
F_{n-1})(f)=\sum_{s=1}^{\gamma(n)}\phi_{ns}
\end{equation}
and
\begin{equation}\label{theoDPTprop1}
\supp(\phi_{ns_1})\cap\supp(\phi_{ns_2})=\emptyset \mbox{ for
}s_1\neq s_2.
\end{equation}
\item[b)] Let $n_1,n_2, s_1,s_2\in\mathbb{N}$ be such that
$n_1<n_2$, $1\leq s_1\leq\gamma(n_1)$ and $1\leq
s_2\leq\gamma(n_2)$. Then
\begin{equation}\label{theoDPTprop2}
\supp(\phi_{n_1s_1})\cap\supp(\phi_{n_2s_2})\neq
\emptyset\Longrightarrow
\supp(\phi_{n_1s_1})\subset\supp(\phi_{n_2s_2})
\end{equation}
\end{itemize}
\end{theorem}
\begin{proof}
a) Denote $g=F_{n-1}(f)$. We have
\begin{equation}\label{theoDPT01}
 ((id-U_n\circ L_n)\circ F_{n-1})(f)=(id-L_n)(g)+(id-U_n)(L_n(g)),
\end{equation}
where the first term in the sum on the right hand side is
nonnegative while the second one is nonpositive.  Let $x\in
\mathbb{Z}^2$ be such that $((id-L_n)(g)>0$. It follows from
Theorem \ref{theoLocMaxMin1} that there exists a local maximum set
$V$ of $g$ such that $x\in V$ and $\card(V)\leq n$. Since
$g=(U_{n-1}\circ L_{n-1})(F_{n-2}(f))$ does not have local maximum
set of size smaller than $n$, see Corollary
\ref{theoLUcompsmooth}, this implies that $\card(V)=n$ and that
$g$ is a constant on $V$. Furthermore, $((id-L_n)(g)(y)=0$ for
$y\in\adj(V)$. Indeed, if $((id-L_n)(g)(y)>0$ for some
$y\in\adj(V)$, then $y$ belongs to a local maximum set $W$ of $g$
and $\card(W)\leq n$. However, any maximum set containing $y$ must
contain $V$ as well which implies $\card(W)\geq n+1$, a
contradiction. In this way we obtain that the support of
$(id-L_n)(g)$ is a union of disjoint connected sets of size $n$,
that is,
\[
\supp((id-L_n)(g))=V_1\cup V_2\cup ...\cup
V_{\gamma_1(n)},
\]
where $V_s\in\mathcal{C}$,
$\card(V_s)=n$, $s=1,...,\gamma_1(n)$ and $V_{s_1}\cap
V_{s_2}=\emptyset$ for $s_1\neq s_2$. Furthermore, $(id-L_n)(g)$
is a constant on each set $V_s$. If $(id-L_n)(g)(x)=\alpha_s$ for
$x\in V_s$, then
\begin{equation}\label{theoDPT02}
(id-L_n)(g)=\sum_{s=1}^{\gamma_1(n)}\phi_{ns}\ ,
\end{equation}
where
\[
\phi_{ns}(x)=\left\{\begin{tabular}{lll}$\alpha_s$&if&$x\in
V_s$\\0&if&$x\in \mathbb{Z}^2\setminus V_s$\end{tabular}\right.
\]
Applying the same approach to the second term in (\ref{theoDPT01})
we obtain
\[
\supp((id-L_n)(g))=W_1\cup W_2\cup ...\cup W_{\gamma_2(n)},\]
where $W_s\in\mathcal{C}$, $\card(W_s)=n$, $s=1,...,\gamma_2(n)$,
$W_{s_1}\cap W_{s_2}=\emptyset$ for $s_1\neq s_2$ and
\begin{equation}\label{theoDPT03}
(id-U_n)(L_n(g))=\sum_{s=\gamma_1(n)+1}^{\gamma(n)}\phi_{ns},
\end{equation}
where $\gamma(n)=\gamma_1(n)+\gamma_2(n)$ and
$\supp(\phi_{ns})=W_{s-\gamma_1(n)}$,
$s=\gamma_1(n)+1,...,\gamma(n)$. Note that $\phi_{ns}$,
$s=1,...,\gamma_1(n)$, are upward (positive) pulses while
$\phi_{ns}$, $s=\gamma_1(n)+1,...,\gamma(n)$ are downward
(negative) pulses. We obtain (\ref{theoDPTprop0}) by substituting
(\ref{theoDPT02}) and (\ref{theoDPT03}) in (\ref{theoDPT01}). It
only remains to show that $V_{s_1}\cap W_{s_2}=\emptyset$ for all
$s_1=1,...,\gamma_1(n)$, $s_2=1,...,\gamma_2(n)$. Indeed, assume
that $V_{s_1}\cap W_{s_2}\neq\emptyset$. From the definition of
the operator $L_n$, there exists $y\in\adj(V_{s_1})$ such that
$L_n(g)(x)=g(y)$ for $x\in V_{s_1}\cap\{y\}$. Therefore,
$V_{s_1}\cap\{y\}\subseteq W_{s_2}$, which implies that
$\card(W_{s_2})\geq n+1$. Since the size of each one of the sets
$W_1,...W_{\gamma_2(n)}$ is $n$, none of them intersects
$V_{s_1}$.

b) Let $\supp(\phi_{n_1s_1})\cap\supp(\phi_{n_2s_2})\neq
\emptyset$. It follows from the construction of
(\ref{theoDPTprop0}) derived in a) that the functions $F_n(f)$ and
$L_{n+1}(F_n(f))$, $n\geq n_1$, are constants on the set
$\supp(\phi_{n_1s_1})$. Furthermore, the set
$\supp(\phi_{n_2s_2})$ is a local maximum set of $F_{n_2-1}(f)$ or
a local minimum set of $L_{n_2}(F_{n_2-1}(f))$. From the
definition of local maximum set and local minimum set it follows
that $\supp(\phi_{n_1s_1})\subset\supp(\phi_{n_2s_2})$.
\end{proof}

Using Theorem \ref{theoDPT}, the equality (\ref{decomp}) can be
written in the form
\begin{equation}\label{DPTm}
f=\sum_{k=1}^m\sum_{s=1}^{\gamma(k)}\phi_{ks}+F_m(f).
\end{equation}
If the function $f$ has finite support, e.g. as in the case of
images, then $F_m(f)$ is a constant for a sufficiently large $m$.
Then we have
\begin{equation}\label{DPT}
f=\sum_{k=1}^m\sum_{s=1}^{\gamma(k)}\phi_{ks}+c,
\end{equation}
where $c=F_m(f)(x)$, $x\in\mathbb{Z}^2$. The equality (\ref{DPT})
is a discrete pulse decomposition of $f$, where the pulses have
the properties (\ref{theoDPTprop1})--(\ref{theoDPTprop2}). It is
generally accepted that an image is perceived through the
contrast, that is, the difference in the luminosity of neighbor
pixels. The discrete pulse transform (\ref{DPT}) extracts all such
differences as single pulses. Hence, (\ref{DPT}) can be a useful
tool in the analysis of images. Since the information in an image
is in the contrast, the total variation of the luminosity function
is an important measure of the quantity of this information. Image
recovery and noise removal via total variation minimization are
discussed in \cite{ChambolleLions} and \cite{RudinOsherFatemi}. It
should be noted that there are several definition of total
variation of functions of multi-dimensional argument (Arzelà
variation, Vitali variation, Pierpont variation, Hardy variation,
etc.). In the applications cited above the total variation is the
$L^1$ norm of a vector norm of the gradient of the function. Here
we consider a discrete analogue of this concept.

\begin{definition}\label{defTV}
Let $f\in\mathcal{A}(\mathbb{Z}^2)$. If
\begin{equation}\label{defTV1}
\sum_{i=-\infty}^\infty\sum_{j=-\infty}^\infty(|f(i+1,j)-f(i,j)|+|f(i,j+1)-f(i,j)|)<\infty
\end{equation}
then $f$ is said to be of bounded variation. The sum on the left
side on the inequality (\ref{defTV1}) is called total variation of
$f$ and is denoted by $TV(f)$.
\end{definition}

As mentioned in the introduction, the LULU operators for sequences
are total variation preserving. We show here that their
two-dimensional counterparts considered in this section have the
same property with respect to the total variation as given in
Definition \ref{defTV}.

Let us denote by $BV(\mathbb{Z}^2)$ the set of all functions of
bounded variation in $\mathcal{A}(\mathbb{Z}^2)$. Clearly, all
functions of finite support are in $BV(\mathbb{Z}^2)$. In
particular, the luminosity functions of images are in
$BV(\mathbb{Z}^2)$. The total variation given in Definition
\ref{defTV} is a semi-norm on $BV(\mathbb{Z}^2)$. In particular,
this implies that
\begin{equation}\label{seminormprop2D}
TV(f+g)\leq TV(f)+TV(g).
\end{equation}
The total preservation property is defined for operators on
$BV(\mathbb{Z}^2)$ as in Definition \ref{defTVP}, where
$\mathbb{Z}$ is replaced by $\mathbb{Z}^2$.

\begin{theorem}\label{theoTVP}
The operators $L_n$, $U_n$, n=1,2,..., and their compositions are
all total variation preserving.
\end{theorem}
\begin{proof} Let $f\in BV(\mathbb{Z}^2)$ and
$(i,j)\in\mathbb{Z}^2$. We will show that
\begin{eqnarray}
|f(i,j)-f(i+1,j)|&=&|L_n(f)(i,j)-L_n(f)(i+1,j)|\nonumber\\
&&+\
|(id-L_n)(f)(i,j)-(id-L_n)(f)(i+1,j)|\label{theoTVP01}\hspace{1cm}
\end{eqnarray}
\underline{Case 1.} $L_n(f)(i,j)<f(i,j)$. In follows from Theorem
\ref{theoLocMaxMin1} that there exists a local maximum set $V$
such that $(i,j)\in V$ and $\card(V)\leq n$. Without loss of
generality we may assume that $V$ is the largest set with the said
properties. Then $L_n(f)(x)=f(z)$, $x\in V$, where $z\in\adj(V)$
is such that $f(z)=\max\limits_{y\in\adj(V)}(f)(y)$. Since
$(i+1,j)$ is a neighbor to $(i,j)$, see (\ref{NeighborCond2}), we
have either $(i+1,j)\in V$ or $(i+1,j)\in\adj(V)$.

\noindent\underline{Case 1.1} $(i+1,j)\in V$. Then
$L_n(f)(i,j)-L_n(f)(i+1,j)=f(z)-f(z)=0$ and (\ref{theoTVP01})
trivially holds.

\noindent\underline{Case 1.2} $(i+1,j)\in \adj(V)$. Then $(i+1,j)$
cannot be element of a local maximum set of size smaller or equal
to $n$. Therefore, $L_n(f)(i+1,j)=f(i+1,j)\leq f(z)=L_n(f)(i,j)$,
which implies (\ref{theoTVP01}).

\noindent\underline{Case 2.} $L_n(f)(i,j)=f(i,j)$. If
$L_n(f)(i+1,j)=f(i+1,j)$ the equality (\ref{theoTVP01}) trivially
holds. If $L_n(f)(i+1,j)<f(i+1,j)$, then we obtain
(\ref{theoTVP01}) by repeating the argument in Case 1.2 where the
points $(i,j)$ and $(i+1,j)$ change places.

Similarly to (\ref{theoTVP01}) we prove that
\begin{eqnarray*}
|f(i,j)-f(i,j+1)|&=&|L_n(f)(i,j)-L_n(f)(i,j+1)|\nonumber\\
&&+\ |(id-L_n)(f)(i,j)-(id-L_n)(f)(i,j+1)|
\end{eqnarray*}
Then by Definition \ref{defTV} we have
\[
TV(f)=TV(L_n(f))+TV((id-L_n)(f)).
\]
The total variation preserving property of $U_n$ is proved in a
similar way.

In order to complete the proof we show that the composition
$A\circ B$ of any two total variation preserving operators $A$ and
$B$ on $BV(\mathbb{Z}^2)$ is also total variation preserving.
Using the total variation preserving property of $A$ and $B$ and
(\ref{seminormprop2D}) we have
\begin{eqnarray*}
TV(f)&=&TV(B(f))+TV((id-B)(f))\\
&=&TV(A(B(f)))+TV((id-A)(B(f)))+TV((id-B)(f))\\
&\geq&TV((A\circ B)(f))+TV(((id-A)\circ B+id-B)(f))\\
&=&TV((A\circ B)(f))+TV((id-A\circ B)(f)).
\end{eqnarray*}
From (\ref{seminormprop2D}) we also obtain $TV(f)\leq TV((A\circ
B)(f))+TV((id-A\circ B)(f))$. Therefore $TV(f)=TV((A\circ
B)(f))+TV((id-A\circ B)(f))$.
\end{proof}

Let function $f\in\mathcal{A}(\mathbb{Z}^2)$ have finite support,
e.g. as in the case of images. Then $f\in BV(\mathbb{Z}^2)$. Using
Theorem \ref{theoTVP} the discrete pulse decomposition (\ref{DPT})
is total variation preserving in the sense that
\begin{equation}\label{DPTTVP}
TV(f)=\sum_{k=1}^m\sum_{s=1}^{\gamma(k)}TV(\phi_{ks}).
\end{equation}
We should remark that representing a function as a sum of pulses
can be done in many different ways. However, in general, such
decompositions increase the total variation, that is, we might
have strict inequality in (\ref{DPTTVP}) instead of equality. The
equality in (\ref{DPTTVP}) means that no additional total
variation, or noise, is created via the decomposition.

\section{Partial reconstructions and noise removal}

\begin{figure}[h]
\begin{center}
\includegraphics[scale=0.25]{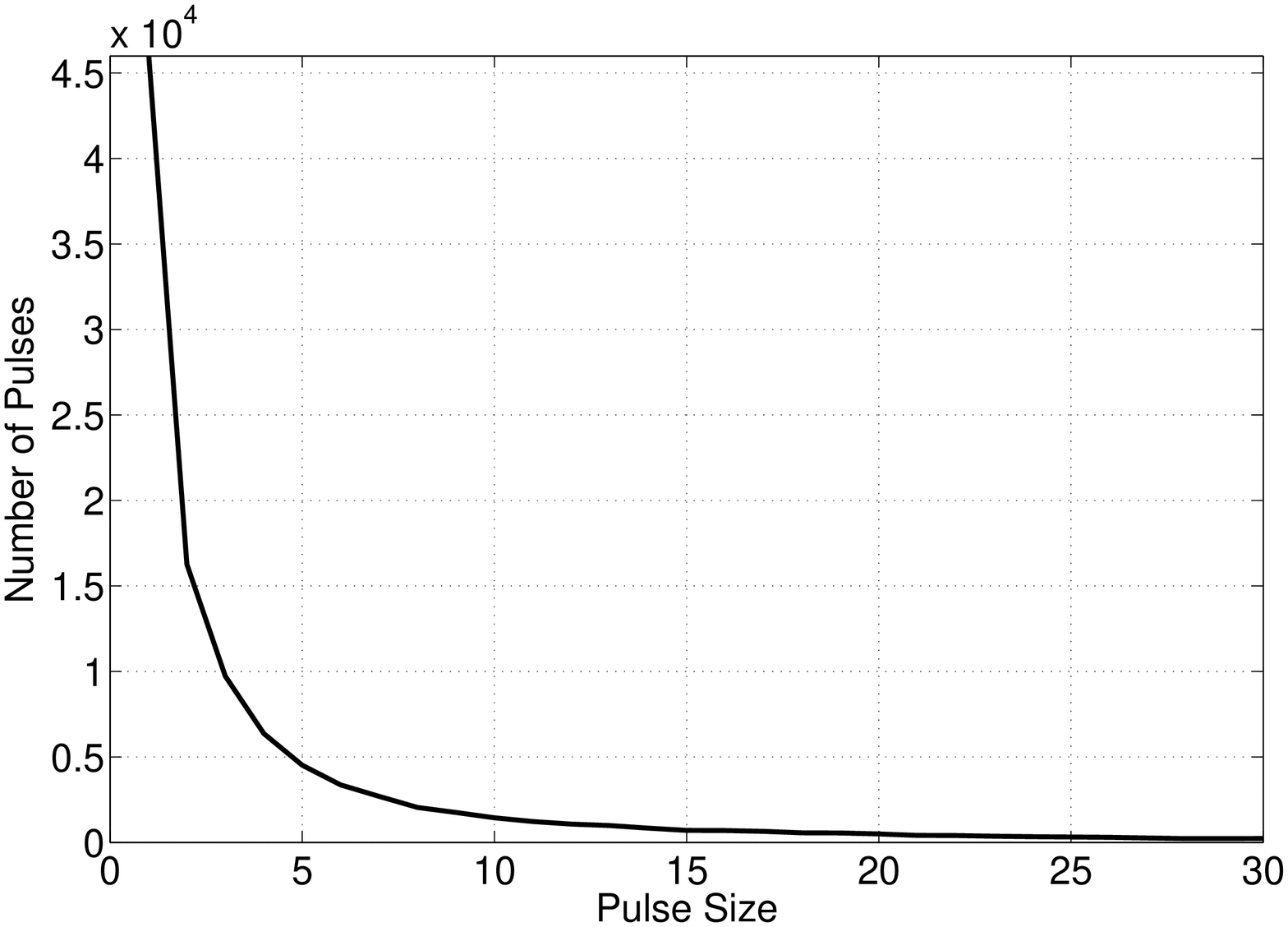}
\includegraphics[scale=0.25]{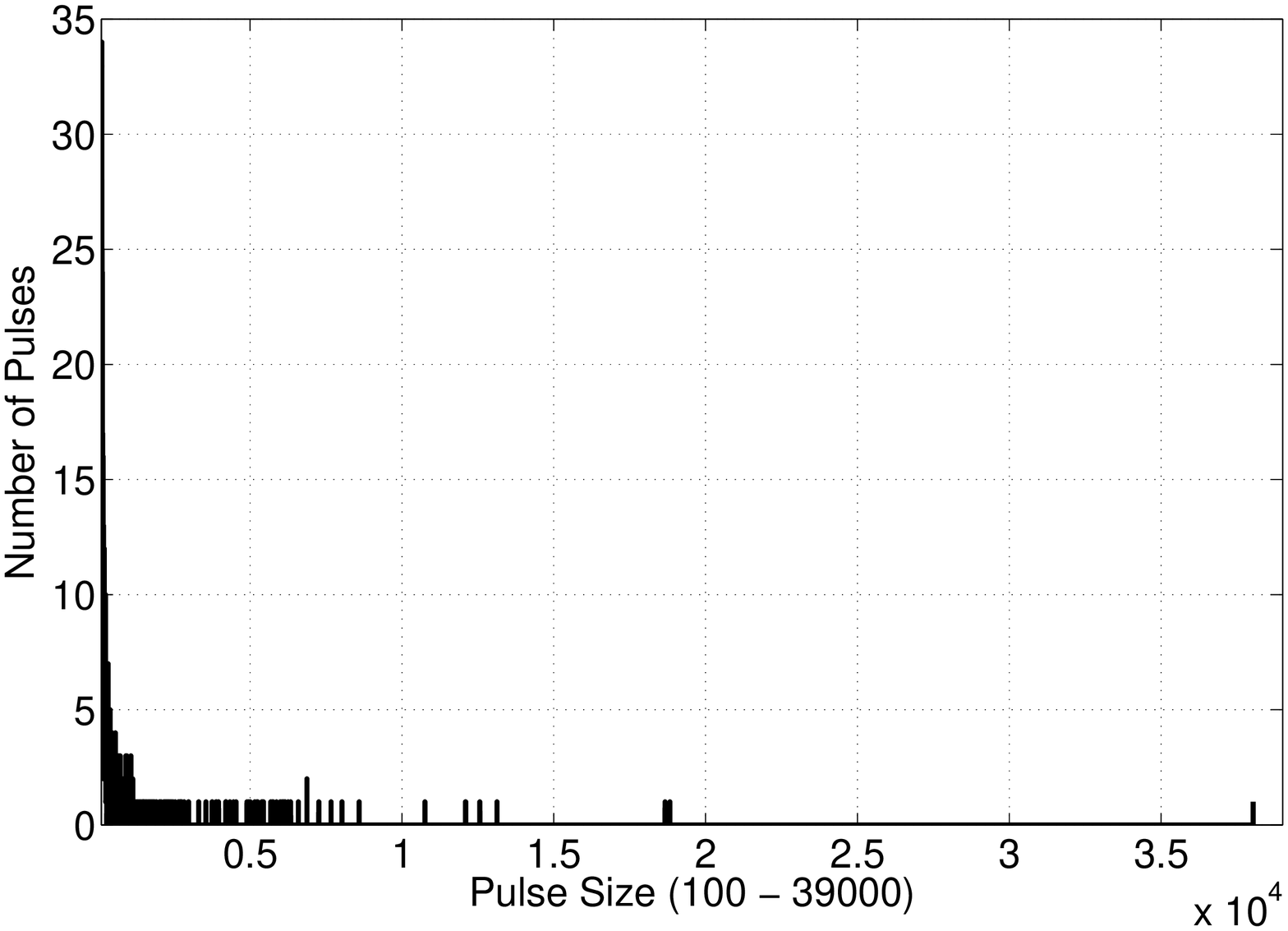}
\caption{Pulse distribution of random noise\label{noisepulsehist}}
\end{center}
\end{figure}

\begin{figure}[h]
\begin{center}
\includegraphics[scale=0.35]{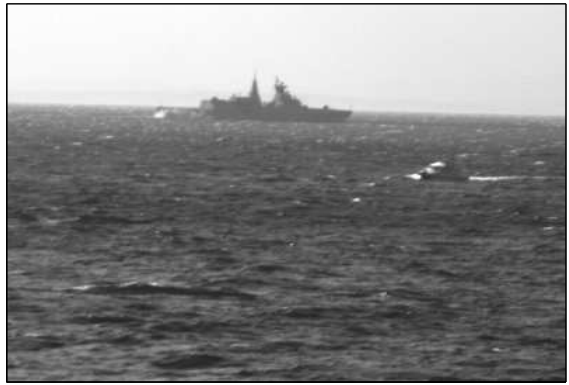}
\caption{A sea image\label{zigzag}}
\end{center}
\end{figure}

\begin{figure}[h]
\begin{center}
\includegraphics[scale=0.25]{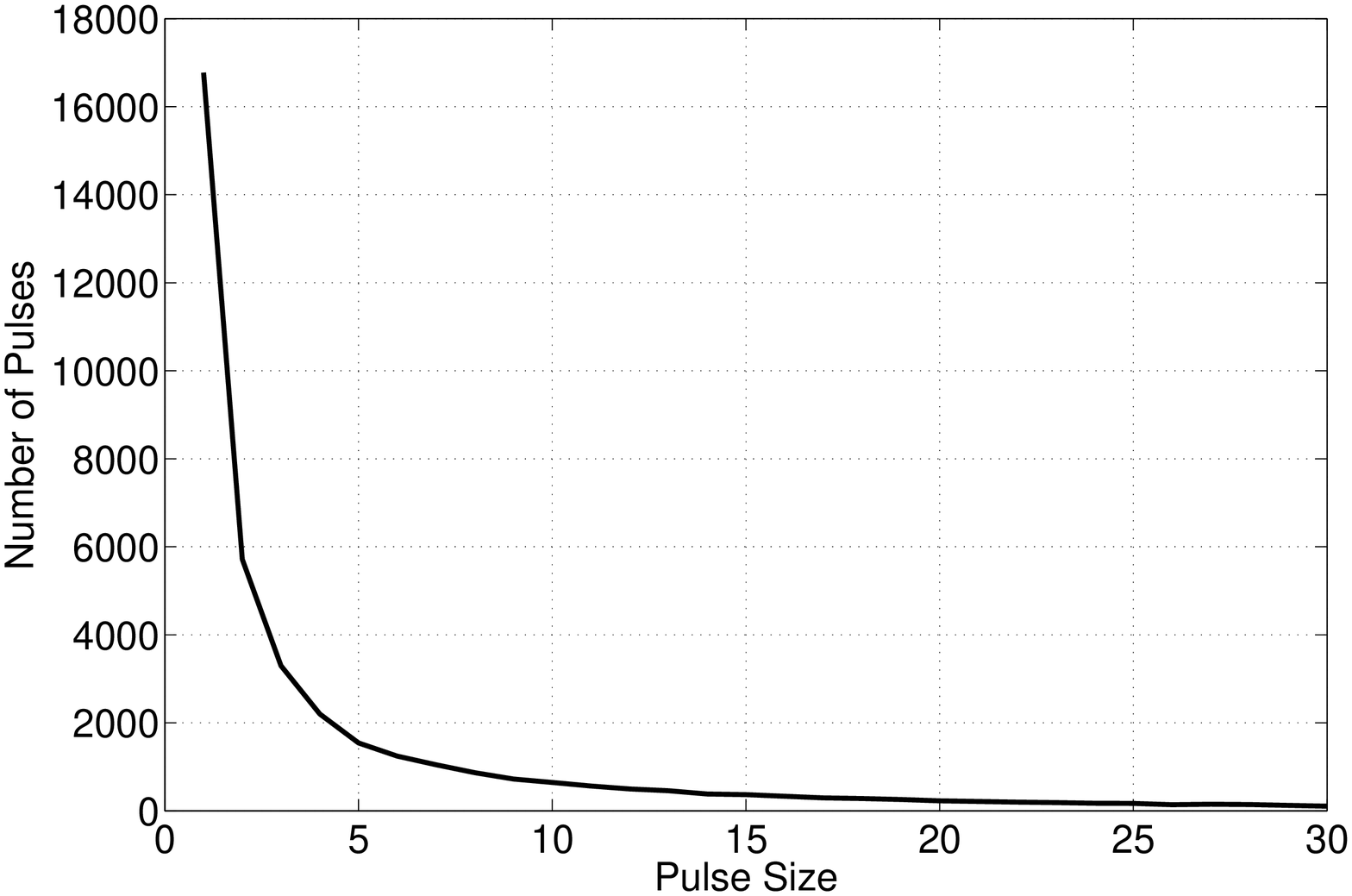}
\includegraphics[scale=0.25]{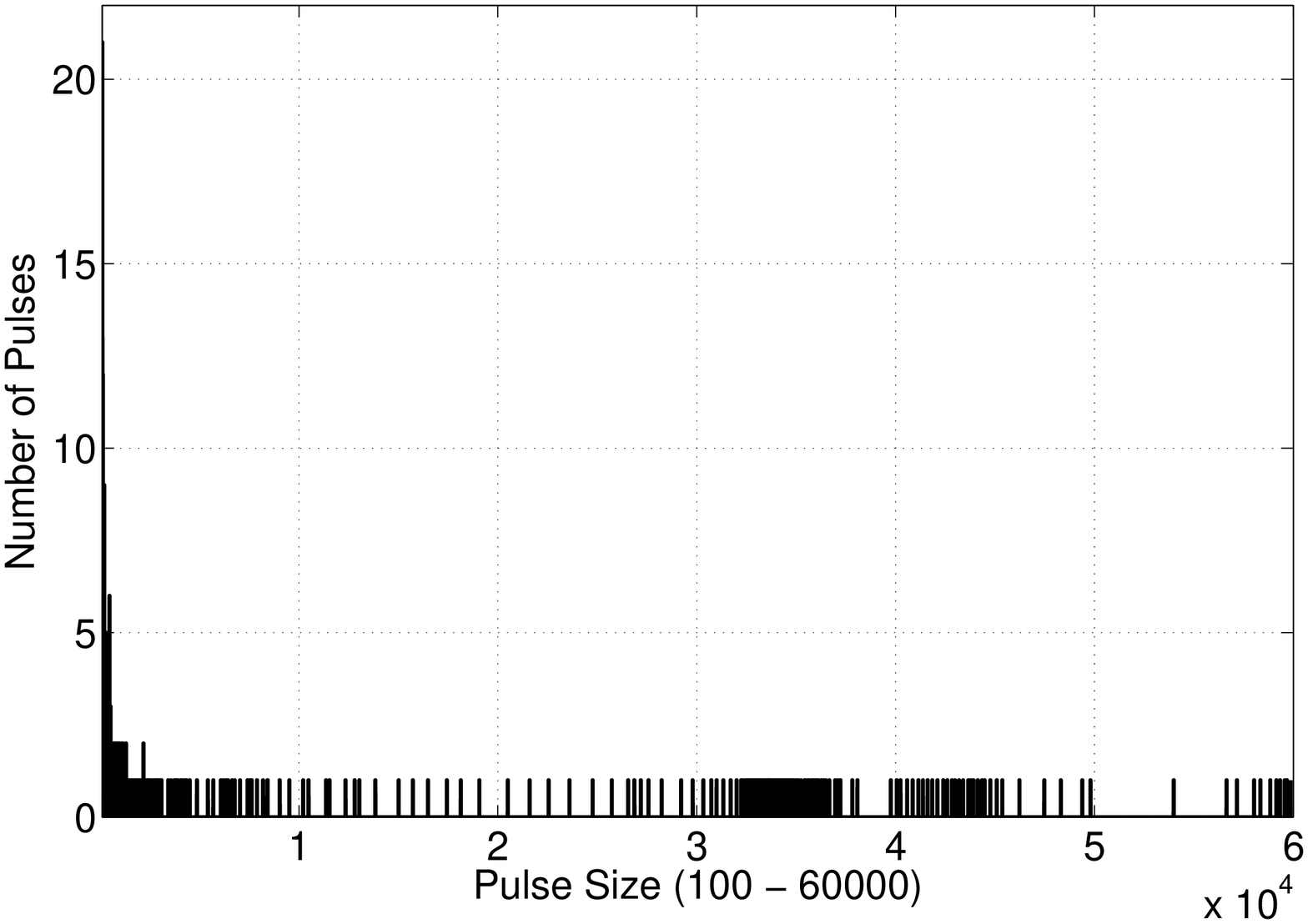}
\caption{Pulse distribution of the sea
image\label{zigzagpulsehist}}
\end{center}
\end{figure}

\begin{figure}[h]
\begin{center}
\noindent\raisebox{3cm}{(a)}\ \ \
\includegraphics[scale=0.35]{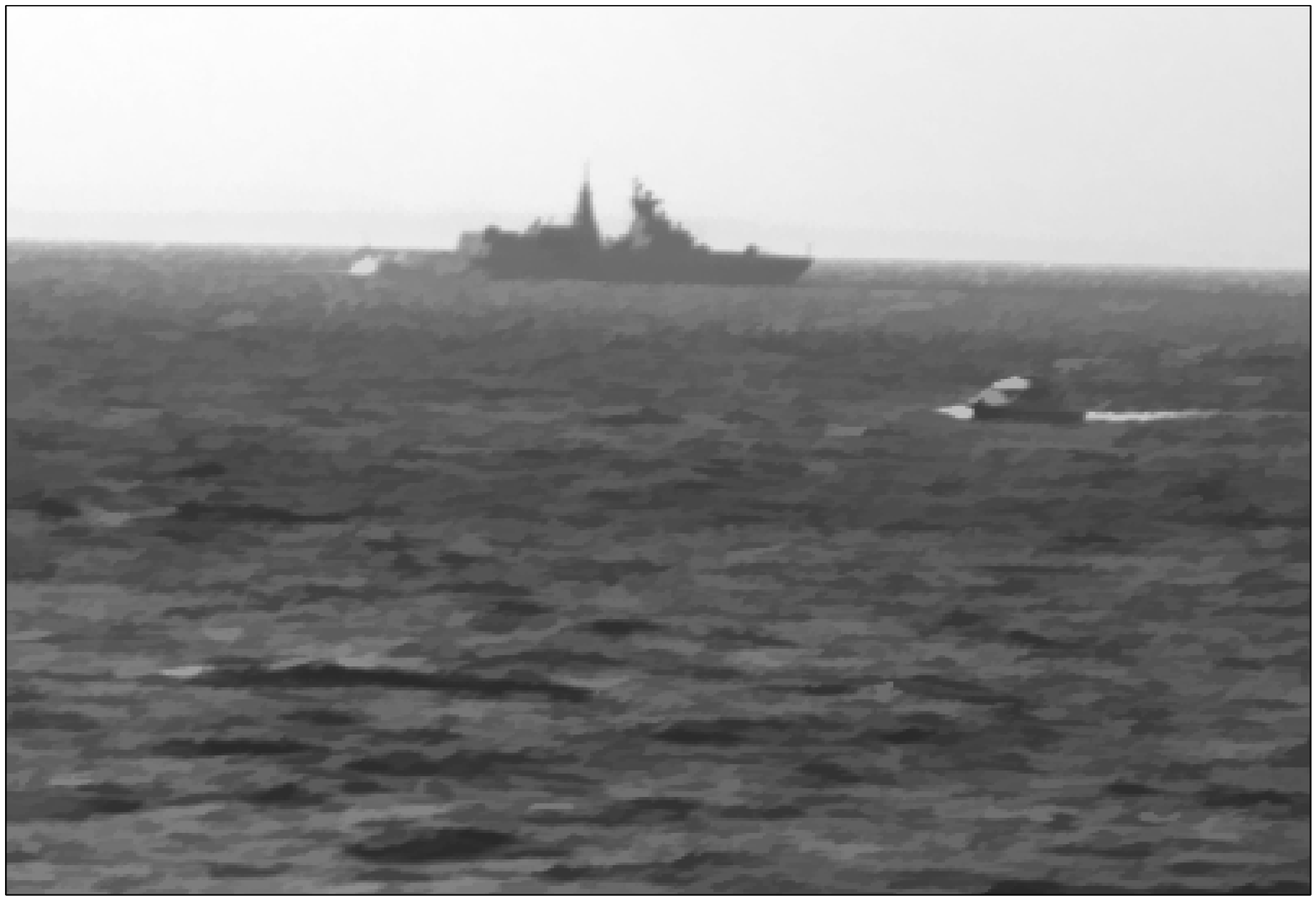}\\
\raisebox{3cm}{(b)}\ \ \
\includegraphics[scale=0.35]{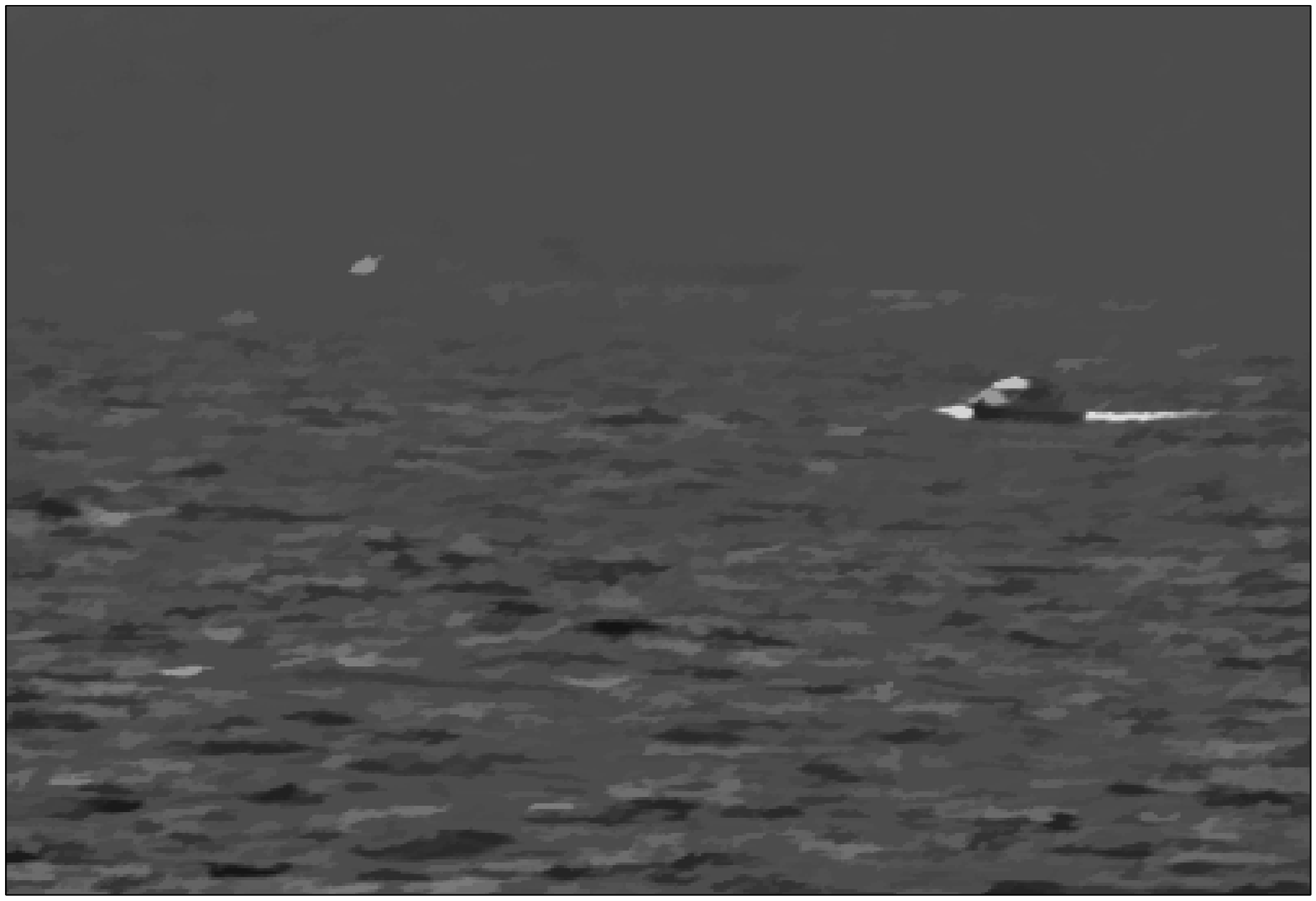}\\
\raisebox{3cm}{(c)}\ \ \
\includegraphics[scale=0.35]{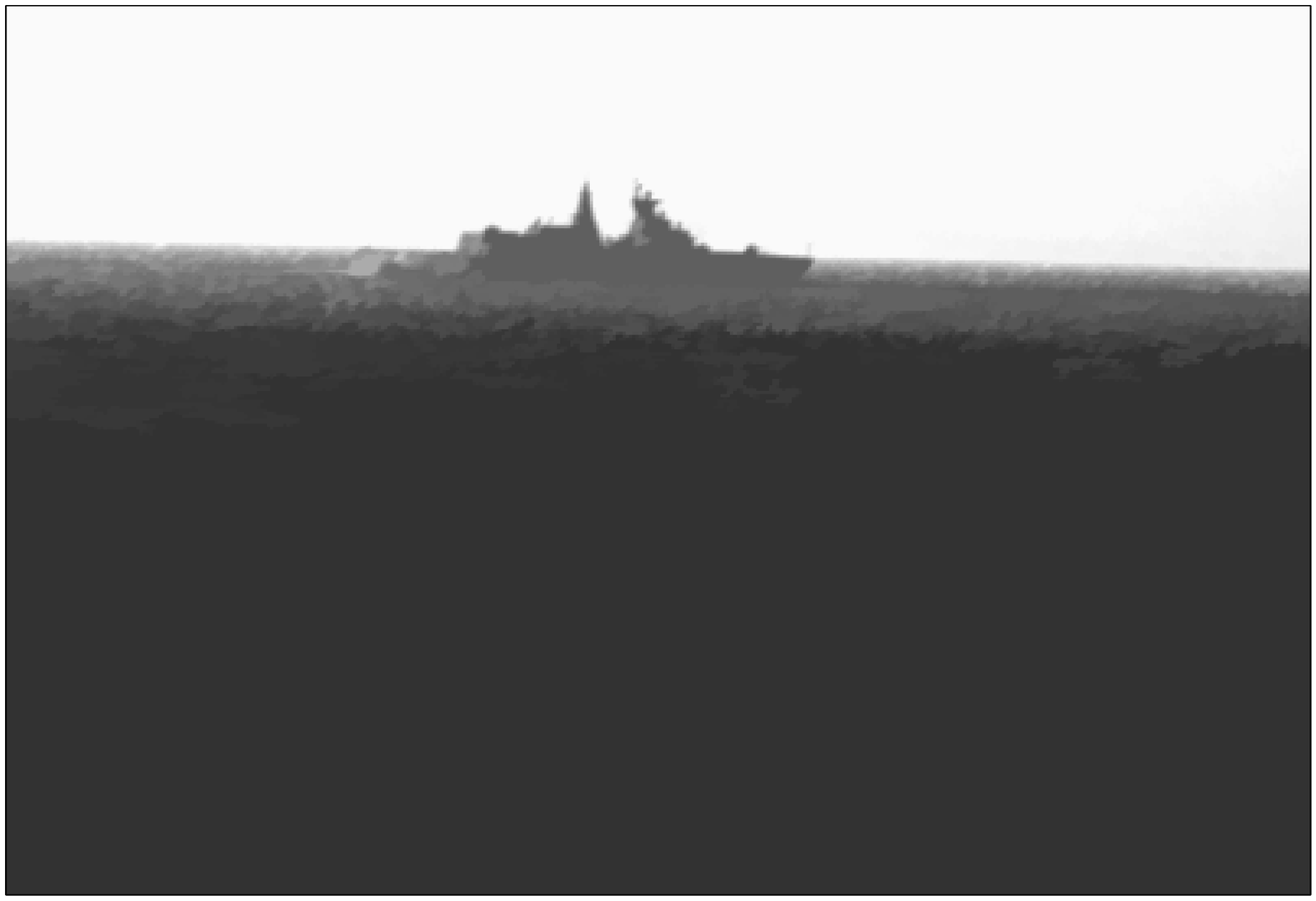}
\caption{Partial reconstructions of the sea image: (a) Pulses of
size larger than 20; (b) Pulses of size from 21 to 400; (c) Pulses
of size from 30000 to 50000.\label{zigzagrec}}
\end{center}
\end{figure}

Possibly the simplest application of the discrete pulse
decomposition (\ref{DPT}) is via partial reconstructions of
images. This can be used for example in removing noise or
extracting features of interest. Random noise has very distinctive
discrete pulse decomposition characterized by fast decrease of the
number of pulses with the increase of the pulse size. The number
of pulses in decomposition (\ref{DPT}) versus their size for a
$300\times 400$ image of random noise (the luminosity at each
pixel is an independent uniformly distributed random variable) is
plotted on Figure \ref{noisepulsehist}. It is apparent that random
noise seldom generates pulses of large size. In fact, 90\% of the
pulses represented on Figure \ref{noisepulsehist} are of size less
than or equal to 20 and only about 2\% have size greater than 100.
Hence by removing the pulse of small support we remove large
portion of any impulsive noise. Figure \ref{zigzagpulsehist} gives
in the same format the pulse distribution of the image on Figure
\ref{zigzag}. A large portion of the pulses has small support but,
unlike Figure \ref{noisepulsehist}, we have also significant
number of pulses with relatively larger support. Partial
reconstruction of the image by using pulses of selected sizes is
given on Figure \ref{zigzagrec}. We can consider (a) as removing
of impulsive noise, (b) as extraction of small features and (c) as
extraction of large features.

\section*{Acknowledgements}

The authors would like to thank Carl Rohwer for his suggestions
and useful discussions.

\end{document}